\documentclass[a4paper]{article}
\usepackage{graphicx}
\usepackage{booktabs} 

\usepackage{enumitem} 
\setlist{leftmargin=6mm}
\usepackage{xspace} 

\usepackage{xcolor}
\usepackage{multirow}
\usepackage{subfig}

\usepackage{wrapfig}

\usepackage{algorithm}
\usepackage{algorithmic}

\usepackage[textsize=tiny]{todonotes}

\usepackage{tikz}
\usepackage{pgfplots}
\usepackage{pgfplotstable}
\usetikzlibrary{arrows}
\usepackage[normalem]{ulem}
\usepackage{lipsum}
\usepackage{multicol}
\usepackage{subfig}
\newcommand{\ignore}[1]{}

\usepackage{graphicx}
\usepackage{caption,epstopdf,url,amsmath}
\usepackage{caption}
\usepackage[sort,numbers]{natbib}
\usepackage{verbatim}

\usepackage[numbers]{natbib}

\usepackage{comment}
\newcommand{\val}{\mbox{\em{val}}}

\title{A Preliminary  Exploration of Floating Point Grammatical Evolution}
\author{Brad Alexander}

\begin{document}
\maketitle

\begin{abstract}
Current GP frameworks are highly effective on a range of real and simulated benchmarks.  However, due to the 
high dimensionality of the genotypes for GP,  the task of  visualising the fitness landscape for GP search can be difficult. This paper describes a new framework: Floating Point Grammatical Evolution (FP-GE) which uses a single floating point genotype to encode an individual program. This encoding permits easier visualisation of the fitness landscape arbitrary problems by providing a way to map fitness against a single dimension. The new framework also makes it trivially easy to apply continuous search algorithms, such as Differential Evolution, to the search problem.  In this work, the FP-GE  framework is  tested against several regression problems, visualising the search landscape for these and comparing different search meta-heuristics. 
\end{abstract}

\section{Introduction}
Genetic programming (GP) has a long history and has been applied, with great effect, in a very wide range of applications\cite{mcphee2008field,poli2014genetic}.
Genotype representations in GP are diverse 
and include, abstract syntax trees\cite{koza1994genetic}, cartesian graphs\cite{miller2000cartesian}, linear code sequences\cite{brameier2007linear}, and sequences of numerical codons\cite{ONeill:2001}, among many others. To date, most of these representations have been discrete and/or multidimensional. 
This use of discrete representations is not unsurprising given that the target phenotype -- program code -- is itself a  discrete structure composed of multiple components. However, these discrete, multidimensional representations have two limitations. First, it is non-trivial to visualise the fitness landscape for most applications of GP\cite{langdon2013foundations}. Second, a discrete representation makes it less easy to apply continuous optimisation 
heuristics to the GP problem.

This work attempts to address these issues by defining a methodology for evolving code with each individual's genotype conists of a single high-precision floating point number. To allow testing of  this approach on a variety 
of target problems and grammars, new framework is based on an existing 
 Grammatical Evolution (GE) implementation. This new approach is labelled Floating-Point Grammatical Evolution (FP-GE).

The contributions of this work are as follows. 
It introduces a simple genotype to phenotype mapping from a floating point number to program code. This mapping is encoded using a choice of depth-first or breadth-first traversal of the target language production grammars. Further, this work introduces corresponding mutation and crossover operators for working with the new genotype. It maps, at different resolutions, the fitness landscape for six different grammars on eight different regression problems and relates this landscape to phenotype complexity.
Moreover, this work applies, a continuous search heuristic, Differential Evolution\cite{storn1997differential} in a more straightforward way than previous works\cite{o2006grammatical,moraglio2010geometric,veenhuis2009tree,moraglio2011geometric,fonlupt2012continuous}. A comparison is made of the performance of the above methods, to standard GE\cite{ONeill:2001}, and to random search in the space of the floating point genotype. The article concludes by  briefly exploring the impact of changing the grammar of an existing problem to improve the search landscape.

The remainder of this paper is structured as follows. The next section outlines related work. Section~\ref{sec:method} outlines the framework and experimental methodology. Section~\ref{sec:results} presents the results of the experiments and, finally,  section~\ref{sec:conc}  concludes and canvasses future work.

\section{Related Work}
There has been substantial work on the use of partial and full  floating point representations in GP\cite{cerny2008using, wu2015deep, o2006grammatical, veenhuis2009tree, medvet2017comparative, moraglio2011geometric, fonlupt2012continuous}. At the most direct level, there has been work using 
continuous optimisation frameworks to evolve floating constants in
new programs\cite{cerny2008using} and tune them in existing programs\cite{wu2015deep}. 

Work aimed at evolving the whole program using floating point encodings\cite{o2006grammatical, veenhuis2009tree, medvet2017comparative, moraglio2011geometric, fonlupt2012continuous} is usually motivated by the desire to apply continuous optimisation frameworks such as Differential Evolution (DE)\cite{storn1997differential}. O'Neill\cite{o2006grammatical}defined GE-DE a form of GE that encodes each codon of the genotype
as a floating point number to be optimised using DE. A similar schema, with different bulk 
operators on the genotype vector, was proposed in~\cite{veenhuis2009tree}. In both these works the genotype
was snapped back to integer values before proceeding with mapping of the genotype to program code. 

Moraglio\cite{moraglio2010geometric, moraglio2011geometric} used DE and downhill search to evolve genotypes based on a continuous metric defined on distances between candidate trees.  A similar approach
with a different metric was used by Kushida\cite{kushida2014novel}. The work presented in this paper differs  in the expression of individuals as directly transcribable from a floating point number. 

In more closely related work, Fonlupt\cite{fonlupt2012continuous} encoded linear GP\cite{brameier2007linear} using a set of four floating point numbers. 
Their work differs from FP-GE both in the cardinality of its representation and the  ability of FP-GE to work with arbitrary grammars. In contrast, to the four dimensions used in the above work, the single floating point genome used in FP-GE makes plotting the fitness landscape relatively straightforward. 

In terms of search frameworks,  this work is based on standard GE~\cite{ONeill:2001}. The GE framework offers flexibility in terms of being able to express programs derived from arbitrary grammars whilst still being effective for program optimisation. It should be noted that there are several recent variants~\cite{medvet2017comparative} that offer better performance through changing the way the genome is mapped to productions. The current article leaves these changes for future work but  does explore breadth-first-search, which is analogous to breadth-first GE\cite{fagan2010analysis}.

\section{Methodology}\label{sec:method}
This section describes the FP-GE framework  and the experimental setup used to test it. 
\subsection{Framework Setup}
The following outlines the FP-GE genotype representation; the genotype to phenotype mapping; fitness function; and the search frameworks  used in the experiments.  All of these are implemented in Python by modifying  the PonyGE 2.0 framework\cite{fenton2017ponyge2} for Grammatical Evolution. 

\paragraph{Genotype}
The genotype for a program in FP-GE is a single high-precision floating point number $\val$. It should be stressed that most floating point representations have a very limited number of significant digits. Such representations would not be effective in this context because they do not encode enough information to represent detailed program code. Instead, the the Python mpmath library\cite{millman2011python} is used to represent the floating-point
representation to a specified level of precision. In this work, a precision level of 150 significant digits is used. The mpmath library overloads all of the common arithmetic operators so every operation on an mpmath number is in the space of these high-precision numbers. In FP-GE the value of $\val$ is always in the range $[0,1]$.

\paragraph{Genotype to Phenotype Mapping}
Extant GE approaches implement a genotype to phenotype mapping which converts a numerical vector
to an individual program, function or code-fragment.  These mappings use the elements of the vector (codons)
to index productions in a user-supplied BNF grammar that defines the set of valid program expressions. 
As the vector is decoded a target abstract syntax tree is populated with nodes specified in the productions
read from the grammar. 

In FP-GE the same basic approach is used but instead of a traversing vector of codons, the single codon: $\val$ is mined  for syntax nodes by the following process.  First the head clause $h$ of the grammar is accessed 
and $\val$ is multiplied by the number of productions $\mbox{\em{numprod}}(h)$ in this clause. This produces a number of the form
$i.r$ where $i$ is a number in the range $0,\ldots,\mbox{\em{numprod}}(h)-1$. The number $i$ is then used 
to index the corresponding production $p$ in the first clause. This production is added to the root node
of the tree. The residual $r$ becomes the new $\val$ for 
the selection of future productions from the child nodes of $p$. 
This process continues either until the tree is complete or some limit on the size of the tree is reached (in this latter case the individual is rejected as invalid).  

In traversing any grammar there is choice of where to look for the next production. This work employed two options. The first is depth-first traversal (DFS), where productions are expanded and $\val$ is transcribed by recursive descent. The second is breadth-first traversal (BFS) where $\val$ expanded into a abstract syntax tree level-by-level. 

\begin{figure}
\centering
\includegraphics[width=0.8\textwidth]{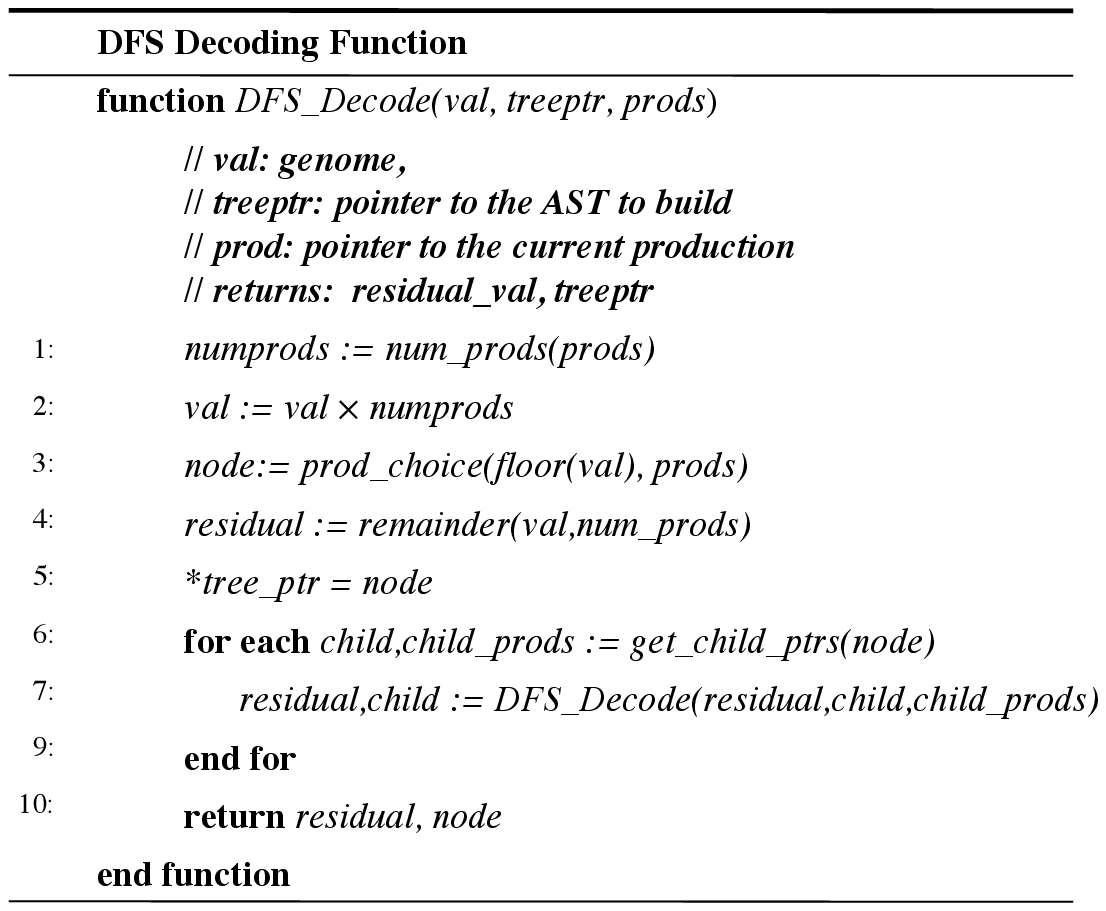}
\caption{\label{alg:dfs} The mapping function for depth-first traversal of a grammar.}
\end{figure}

The pseudo-code for the {\em{DFS\_Decode}} function for mapping $\val$ into a tree in DFS order is shown in Fig.~\ref{alg:dfs}.
This function takes, as parameters, $\val$, {\em{treeptr}} -- a pointer to place to build the next part of the AST and {\em{prods}} the set of choices in the current clause in the target grammar. 
Lines 1 to 4 of the algorithm select the desired production and the new residual values.  Line 5 populates the current node of the tree. The loop in lines 6 to 9 recursively calls {\em{DFS\_Decode}} on the child nodes of the current production (if any).  In performing the DFS 
decoding, the most significant digits in $\val$ will determine the content of the left-most branches of the AST. In contrast, the content of the right-most branches will change radically with very small changes in $\val$ because these branches are decoded last. A-priori DFS decoding will perform better when the choices of left-most nodes in the tree are a stronger determinate of program performance than the right-most nodes. 

\begin{figure}
\centering
\includegraphics[width=0.8\textwidth]{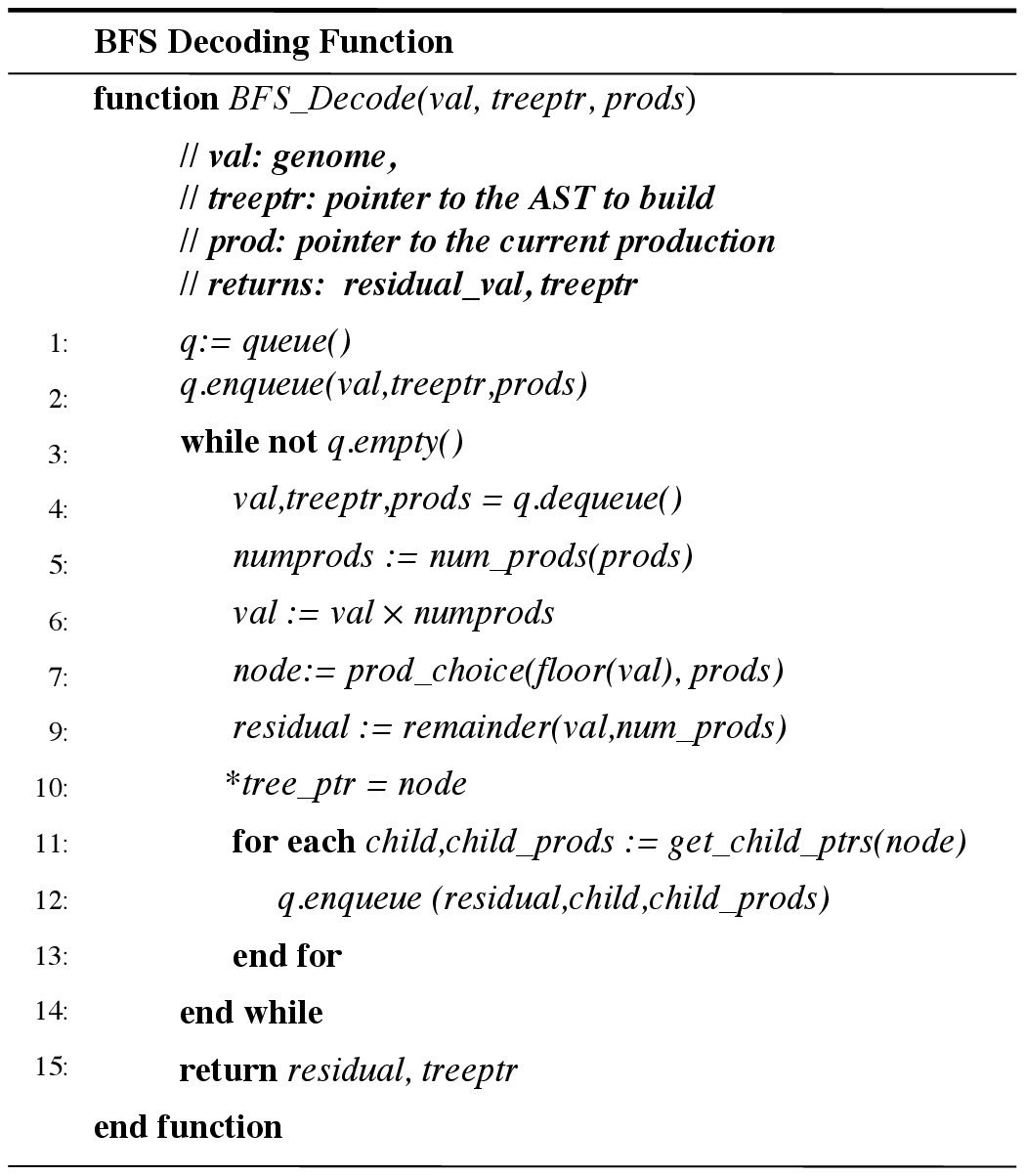}
\caption{\label{alg:bfs} The mapping function for Breadth-first traversal of a grammar.}
\end{figure}

The pseudo-code for the {\em{BFS\_Decode}} function for mapping $\val$ is given in Fig.~\ref{alg:bfs}.
This function takes the same parameters as {\em{DFS\_Decode}} but, instead of traversing the grammar by recursive descent, it uses a queue to drive traversal level-by-level. Lines 1 and 2 set up the queue with a first entry containing the three parameters. Lines 4-9, dequeues the next inputs to work on and selects the next production and works out the new $\val$ and residual. Line 10 populates the current node of the tree. The inner-loop on lines 11 to 13 adds the child nodes in succession to the queue. The outer loop will iterate until the tree is complete or some constraint on the size of the tree is violated (in which case the individual is marked as invalid).  

In performing BFS Decoding, the most significant digits in $\val$ will determine the content of the left-most, top-most nodes of the tree. In contrast, the content of the bottom-most branches will change radically with very small changes in $\val$ because these branches are decoded last. A-priori, BFS Decoding will perform better when the choice of top level nodes in the program tree is a stronger determinate of program performance than the choices of lower levels. 
\paragraph{Fitness Function}
The fitness function converts the AST produced by the mapping 
process into Python code. This code is then run against a set of 
test cases to determine how well the code performs. This work focuses solely on {\em{symbolic regression}}\cite{kotanchek2010symbolic} problem benchmarks distributed with the PonyGE 2.0 implementation. With symbolic regression problems, the program code represents a mathematical function $f$ and the fitness is determined by how well $f$ fits a data set -- with a lower average error being indicative a fitter individual function. 
\paragraph{Search Frameworks}
The search frameworks used in this work are:
\begin{description}
\item{\bf{FP-GE}} A variant of standard GE using the genotype-to-phenotype encodings described above and specialised crossover and mutation operators (described below).
\item{\bf{DE}} Differential Evolution applied to the optimisation $\val$ to produce fit function variants. 
\item{\bf{Random Search}} Random search for fit function variants encoded by $\val$. 
\end{description}
All of the above three search frameworks are applied to the landscapes produced for each grammar by DFS and BFS decoding of $\val$. This gives six search variants. 
In addition, as a basis for comparison, integer GE (int) is also run on all the benchmarks using its standard integer-per-codon encoding. 

\paragraph{Specialised Operators}
In FP-GE, the floating point representation of $\val$ demands specialised mutation and crossover operators. Mutation is implemented as random perturbation selected from a uniform distribution in the range $[-0.05,0.05)$. If a mutation brings $\val$ outside of the range $[0.0,1.0]$ then the value is wrapped to the other side of the range.  The choice of perturbation range was tuned experimentally according to what appeared to be most productive in preliminary experiments. 

Crossover of two individuals $a$ and $b$, where $a<b$ is done by randomly selecting a new value $c\in [a,b]$. As an aside, early experiments where crossover simply bisected $a$ and $b$ appeared to converge prematurely. 

\subsection{Experimental Setup}
\begin{figure}
\centering
\includegraphics[width=0.95\textwidth]{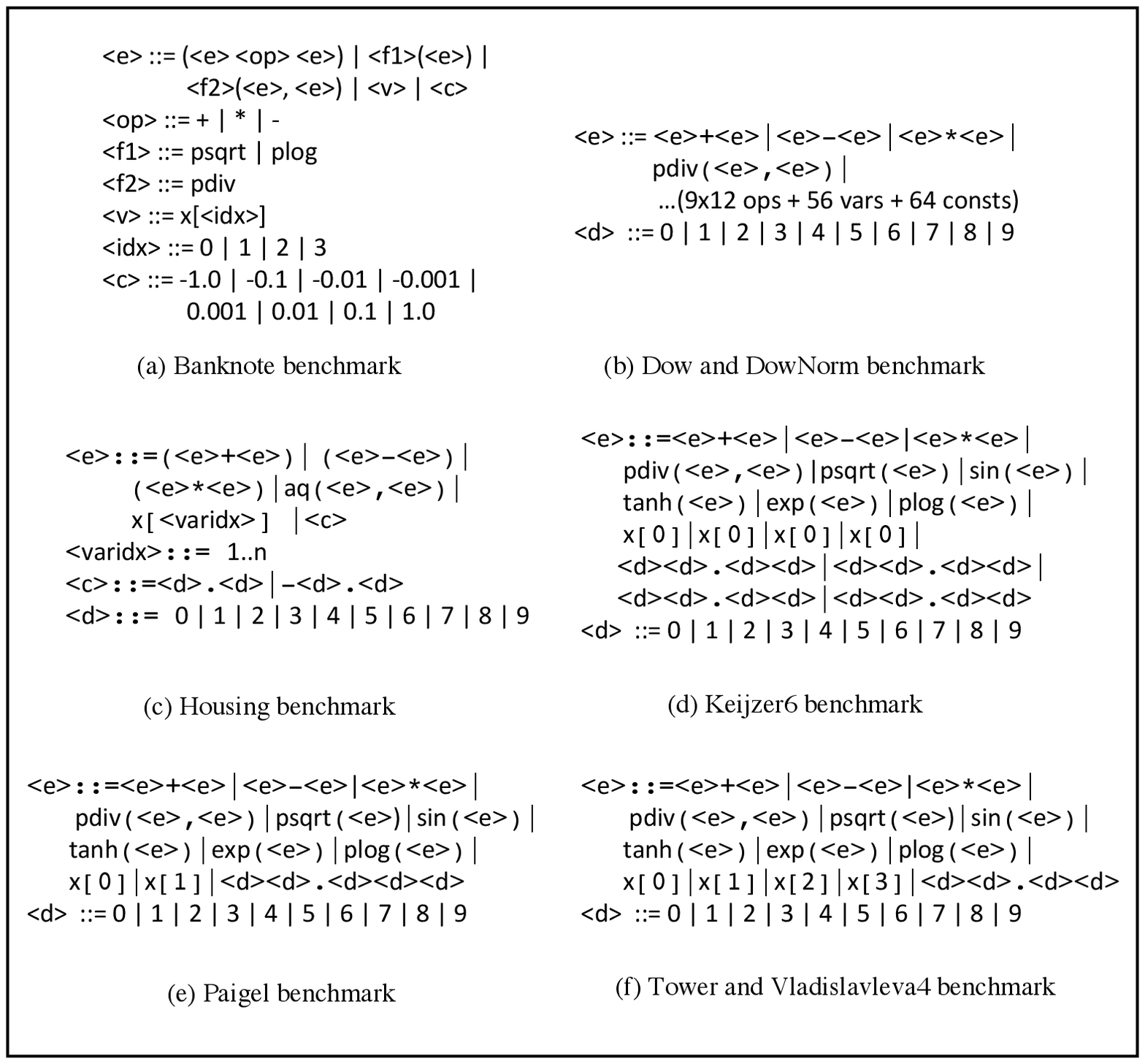}
\caption{\label{fig:grammars} Grammars used in the eight 
benchmarks in this study (note some grammars are used by more than one benchmark).}
\end{figure}

The benchmarks used and the experimental settings used in the 
experiments in the rest of the paper are described below. 
\paragraph{Benchmarks}
The following experiments used eight symbolic regression benchmarks included with the PonyGE implementation. 
These are:
\begin{description}
\item{\bf{Banknote}} Classification problem for features from genuine and forged banknote images. 
\item{{\bf{Dow}}} A regression to data predicting the Dow Jones stock index as a function of 54 variables. 
\item{{\bf{DowNorm}}} A normalised version of the Dow problem. 
\item{{\bf{Housing}}} Boston Housing Data - Price Prediction
\item{\bf{Keijzer6}} Function defined: $f(x) = (30 * x * z) / ((x - 10) * y^2)$
\item{\bf{Paige 1}} Function defined: $f(x,y)=1/(1+x^{-4})+1/(1+y^{-4})$
\item{\bf{Tower}} Tower Data set from Dow Chemical. 
\item{\bf{Vladislavleva-4}} Function defined: $10/(5+\sum_{i=1}^5 (x_i-3)^2)$
\end{description}
In each case, the aim is to minimise the error in the fit 
of the generated function to the benchmark data. 
These benchmarks are reasonably diverse -- ranging from mathematical functions to data harvested by real systems. The number of variables and constants in each problem ranges from single variable to more than 50 for the Dow problems. The grammars used for these benchmarks are shown in Fig.~\ref{fig:grammars}
\paragraph{Experimental Settings}
The experiments were divided into two categories: landscape experiments, and search experiments. The settings
for each, in turn, are described below.
\paragraph{Landscape experiments}
For the landscape experiments for each benchmark, a scan was made  
through $25,000$ values of $\val$ in the range $[0,1.0]$ and evaluated the fitness of the mathematical function generated by each genotype value. In order to avoid systematic noise generated by the regular increments in this range, the scan was started at a small random positive offset from zero. For each benchmark scans were conducted using both {\em{DFS\_decode}} and {\em{BFS\_decode}} because these each generate different landscapes. Data was also collected for the number of nodes encoded in the tree for each value for $\val$. This node-count data provides a proxy for the complexity of the derived tree. 

\paragraph{Search Experiments}
For each benchmark a trace was made of the performance of search for: Integer GE (int); FP-GE DFS (DFS); FP-GE BFS (BFS); Differential Evolution BFS (DE-BFS); Differential Evolution DFS (DE-DFS); Random Search DFS (Rand DFS); Random Search BFS (Rand BFS).

Each experiment was run 30 times and the fitness traces of the best individuals were averaged. Every experimental run was limited to $25,000$ evaluations. For int, DFS, and BFS this was composed of 50 generations with a population of 500. For the DE experiments the population was reduced to 250 and the best results traced continuously over the $25,000$ evaluations. For random search, the best individual was tracked continuously over the $25,000$ evaluations. 

For the integer GE experiments the standard initialisation, mutation, and one-point crossover operators were used and trees were limited to 200 codons. For the FP-GE experiments, trees were limited by depth (depth in the range 14-17) depending on the benchmark. 

\section{Results}\label{sec:results}
The following results are divided into landscape analysis and search results. This is followed by an additional section describing the impact of changing a grammar to make it more easily searchable by FP-GE operators. 
\subsection{Landscape analysis}
\begin{figure}
\centering
\includegraphics[width=0.95\textwidth]{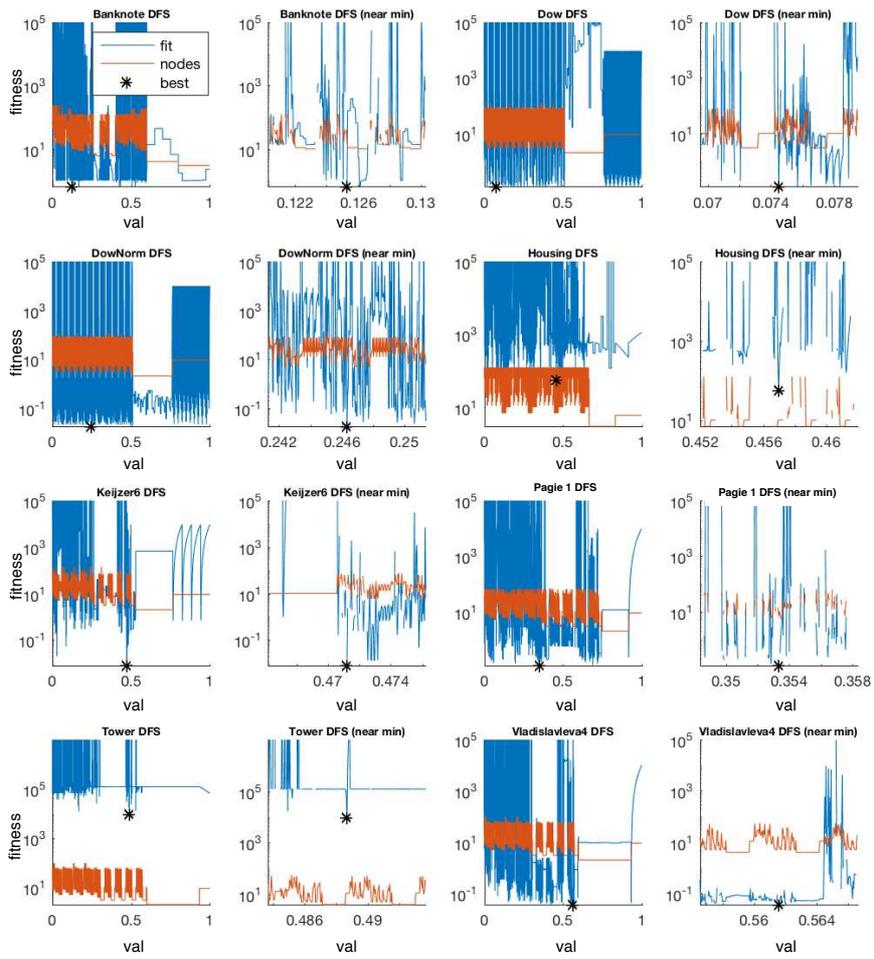}
\caption{\label{fig:landdfs} DFS fitness and complexity landscapes for the eight benchmarks.}
\end{figure}

Figure~\ref{fig:landdfs} shows the landscape of DFS search applied to the benchmarks. 
In the figure, each benchmark is represented by two graphs. 
The ordering of the benchmarks is alphabetical from left to right and top to bottom. 
On the left is a graph showing the {\em{global}} landscape for the benchmark. This traces fitness (in blue) and expression complexity (in red) across the whole scan of $\val$ for the interval $[0.0,1.0]$. The most-fit sampled value in this range
is marked with a black asterisk. Immediately to right of these global graphs are other graphs that sample the 250 values surrounding this most-fit value. 

The first observation that can be made is that the global 
graphs vary from very rough in the left of the range to 
quite smooth on the right. This appears to be an artefact of the grammar structure where the recursive components are first in the grammar and the terminals tend to be last. These left-hand productions also tend to produce the biggest and most varied trees. The relative size of these individual can be observed by the preponderance of larger individuals on the left of the graphs.  Another observation to make is that the graphs surrounding the minimum values are quite rough. For some benchmarks there are discontinuities where the size limits on trees were violated. This indicates a hostile local search landscape near these minimas. 

\begin{figure}
\centering
\includegraphics[width=0.95\textwidth]{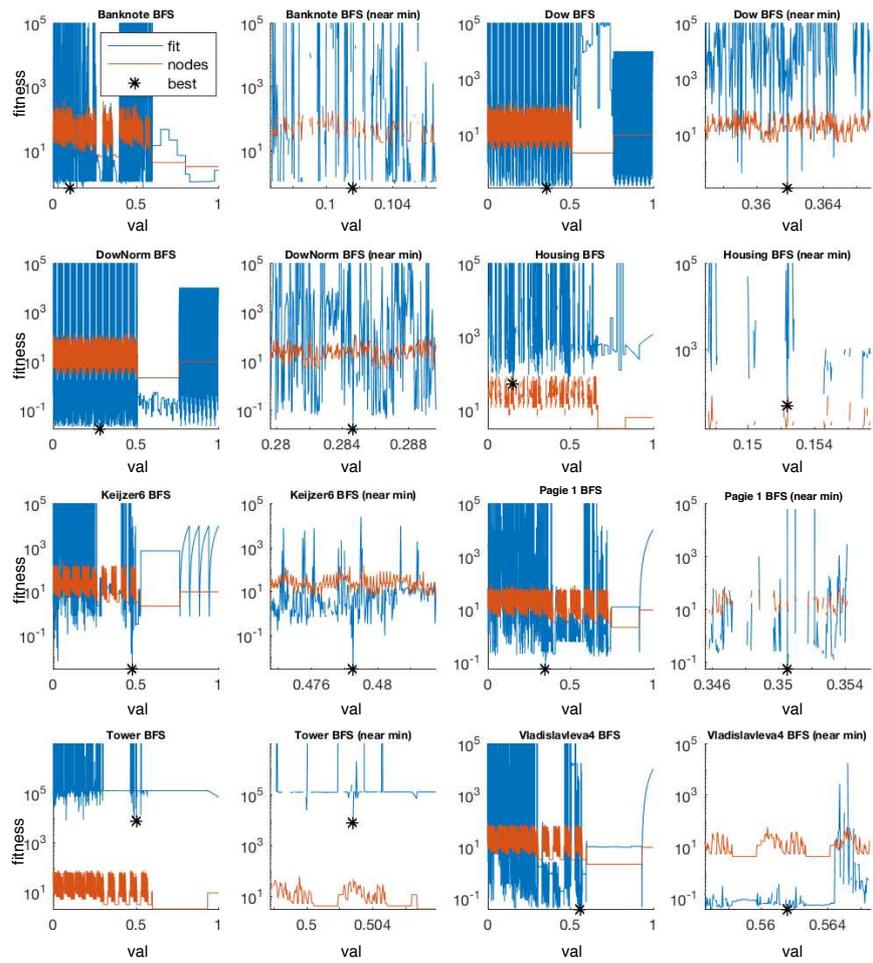}
\caption{\label{fig:landbfs} BFS fitness and complexity landscapes for the eight benchmarks.}
\end{figure}

Figure~\ref{fig:landbfs} shows the corresponding landscapes for a BFS mapping. 
Figure~\ref{fig:landdfs}
At a macro scale, these graphs resemble those for DFS with the more complex derivations and associated noise concentrated on the left. However, the position of the best value is different and the local landscape vary from extremely sparse and noisy, (e.g. the Housing benchmark), to reasonably smooth, e.g. the (Vladislavleva4 benchmark). A-priori this would indicate that the BFS encoding will give better results than DFS on these smoother landscapes. 

\subsection{Search Results}
\begin{figure}
\centering
\includegraphics[width=0.99\textwidth]{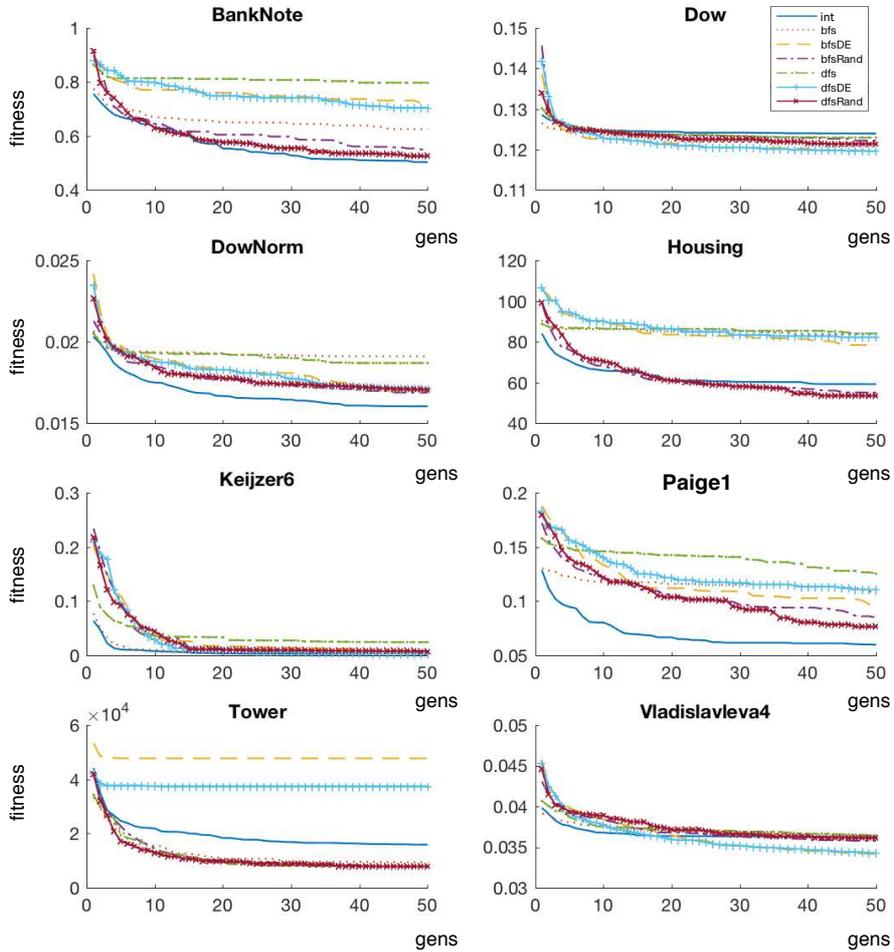}
\caption{\label{fig:search} Search results for the benchmark problems}
\end{figure}

The search results for the seven search algorithms on the 
eight benchmarks are shown in Fig.~\ref{fig:search}
The results are highly varied for each benchmark but a general pattern can be discerned. The benchmarks with poor landscapes for BFS around the minimum value tend to exhibit poor performances for the informed FP-GE search heuristics.
In contrast, the benchmarks with better landscapes tend to have better response to the informed heuristics. Also of note is the relatively good performance of integer GE on some problems, this indicates that for some problems at least, standard GE exhibits a more benign landscape. 
Interestingly, BFS and DFS random search exhibits good performance on a number of benchmarks and outperforms integer GE on the Tower benchmarks. This indicates that the search spaces generated from the grammars, as specified, are not strongly informative. 

\subsection{Changing the Landscape}
One possible cause of the poor landscapes for problems such as the housing data is the structure of the grammar. The first clause of the grammar for the Housing benchmark in Fig.~\ref{fig:grammars}(c) is dominated by binary recursive expressions. This will tend to produce a lot of large trees as this grammar is traversed. This bias can be easily reduced by wrapping up the non-terminal productions in the first clause into a single production that is then elaborated on the next line. This means that at least two-thirds of the time a terminal production is chosen -- reducing the average size of the trees. 

\begin{figure}
\centering
\includegraphics[width=\textwidth]{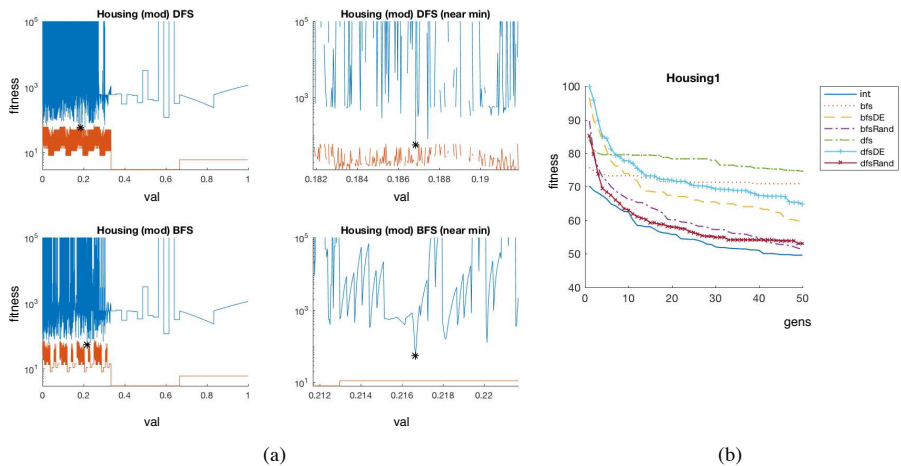}
\caption{\label{fig:house} Landscape for the new grammar for the Housing benchmark.}
\end{figure}

To test the effect of this change the housing experiment was run again with the new grammar. The impact on the landscape is shown in Fig.~\ref{fig:house}(a).
Figure~\ref{fig:landdfs}
The resulting landscapes are somewhat more regular at the global scale. However, near the best-fitness value, the DFS landscape is quite fragmented, this could be due to a preponderance of large trees derived in this neighbourhood. In contrast, the neighbourhood for the minimum BFS values is almost continuous. A-priori this would indicate a more benign search environment.  

This hypothesis about a better search environment for BFS encoding  is supported, to an extent, by the trace data in 
Fig.~\ref{fig:house} (b). 
The most noticeable change from the first trace is the improved performance of the more informed heuristics such as DE. Interestingly the optimisation results appear slightly better overall and the performance of integer GE is better than in the original run. This perhaps indicates that the modified grammar is also producing a better landscape for the GE search framework.

\section{Conclusions}\label{sec:conc}
This paper has demonstrated a new floating-point encoding for grammatical evolution. This encoding has allowed, for the first time,  easy visualisation of fitness landscapes across the range of genotypes. Preliminary indications are that the landscapes produced by some grammars on some problems are quite rough. At this stage the experiments have only sampled at a moderate resolution but it would be interesting to see if this effect is even more visible at a finer-grain of sampling. It would also be interesting to explore the space of grammar design to identify feature of grammars best suited to this encoding. 

In terms of performance, there are some indications that FP-GE can perform comparably to integer GE. The observation that random search performs well in this space aligns with other 
recent observations for standard GE encodings. It would be interesting to identify features of problem definitions for which more informed search heuristics perform better.

\bibliographystyle{acm}
\bibliography{part}

\end{document}